\documentclass[preprint,10pt]{elsarticle}
\usepackage{amssymb}
\usepackage{amsmath}
\usepackage{multirow}
\usepackage{booktabs}
\usepackage{subcaption}
\usepackage{amsfonts}
\usepackage{makecell}
\usepackage{color}

\journal{Pattern Recognition}

\begin{document}

\begin{frontmatter}

\title{
% Multitask and Multimodal Neural Tuning for Large Models 
% \protect\\
One Framework to Rule Them All: Unifying Multimodal Tasks with LLM Neural-Tuning
% \protect\\
% Unified Multimodal Framework: Better Tasks, Smarter Models, Easier Tuning
}

\author[label1,label2]{Hao Sun}
\author[label2]{Yu Song}
\author[label2]{Jiaqing Liu}
\author[label2]{Jihong Hu}
\author[label2]{Yen-Wei Chen\corref{cor1}}
\author[label1]{Lanfen Lin}
\cortext[cor1]{Corresponding Authors}
\affiliation[label1]{organization={College of Computer Science and Technology, Zhejiang University},
            city={Hangzhou},
            state={Zhejiang},
            country={China}}
        
\affiliation[label2]{organization={College of Information Science and Engineering, Ritsumeikan University},
            city={Ibaraki},
            state={Osaka},
            country={Japan}}

\begin{abstract}
Large-scale models have exhibited remarkable capabilities across diverse domains, including automated medical services and intelligent customer support. However, as most large models are trained on single-modality corpora, enabling them to effectively process and understand multimodal signals remains a significant challenge. Current research often focuses on designing task-specific or scenario-specific tuning strategies, which limits the scalability and versatility. To address this limitation, we propose a unified framework that concurrently handles multiple tasks and modalities. In this framework, all modalities and tasks are represented as unified tokens and trained using a single, consistent approach. To enable efficient multitask processing, we introduce a novel tuning strategy termed neural tuning, inspired by the concept of sparse distributed representation in the human brain, where only specific subsets of neurons are activated for each task. Furthermore, to advance research in multimodal and multitask learning, we present a new benchmark, MMUD, which includes samples annotated with multiple task labels spanning reasoning segmentation, referring segmentation, image captioning, and text-to-image generation. By applying neural tuning to pretrained large models on the MMUD benchmark, we demonstrate the ability to handle multiple tasks simultaneously in a streamlined and efficient manner. All codes and datasets will be released at https://github.com/kiva12138/NeuralTuning.
\end{abstract}

\begin{graphicalabstract}
\includegraphics[width=1.0\textwidth]{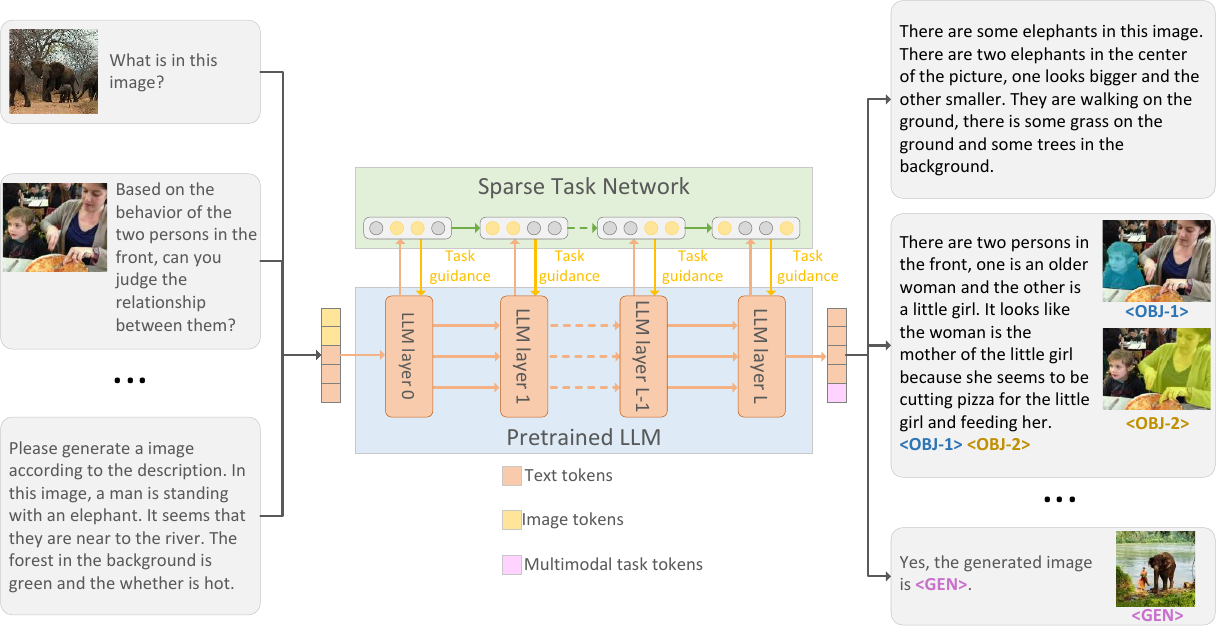}
\end{graphicalabstract}

\begin{highlights}
\item Propose a framework that unifies multimodal tasks with a concise all-in-token manner.
\item The framework enhances the flexibility and scalability of LLMs.
\item Introduce neural tuning, a efficient yet effective task-specific tuning strategy. 
\item Present MMUD, a multimodal dataset for tasks like referring segmentation and generation.  
\item Reach state-of-the-art in performance by involving neural tuning on MMUD.
\end{highlights}

%% Keywords

\end{frontmatter}


\section{Introduction}
Recently, the rapid advancements in deep learning and hardware computing power have propelled the development of large-scale models, achieving significant breakthroughs in applications such as intelligent customer support and autonomous driving. The remarkable success of these models can be attributed to their large-scale architectures and extensive training data, which substantially enhance their contextual understanding and complex reasoning capabilities. However, most large-scale models are pretrained on single-modality corpora due to the challenges associated with acquiring large-scale multimodal datasets and the computational limitations of current hardware. To address this, recent research has explored enabling large language models (LLMs) to process multimodal data through fine-tuning. While these methods improve performance on specific tasks, they predominantly rely on task-specific architectures or tuning strategies (such as referring segmentation and image-text classification~\cite{liang2022expanding,ren2023pixellm,ho2020denoising}), which significantly hinders their scalability and versatility in handling multiple tasks concurrently. Although expanding the capabilities of LLMs through more intricate architectures or tuning strategies is possible, this approach imposes considerable complexity and overhead, limiting the feasibility of extending such models to accommodate additional tasks or datasets.

\begin{figure*}[t]
\centering
\includegraphics[width=1.0\textwidth]{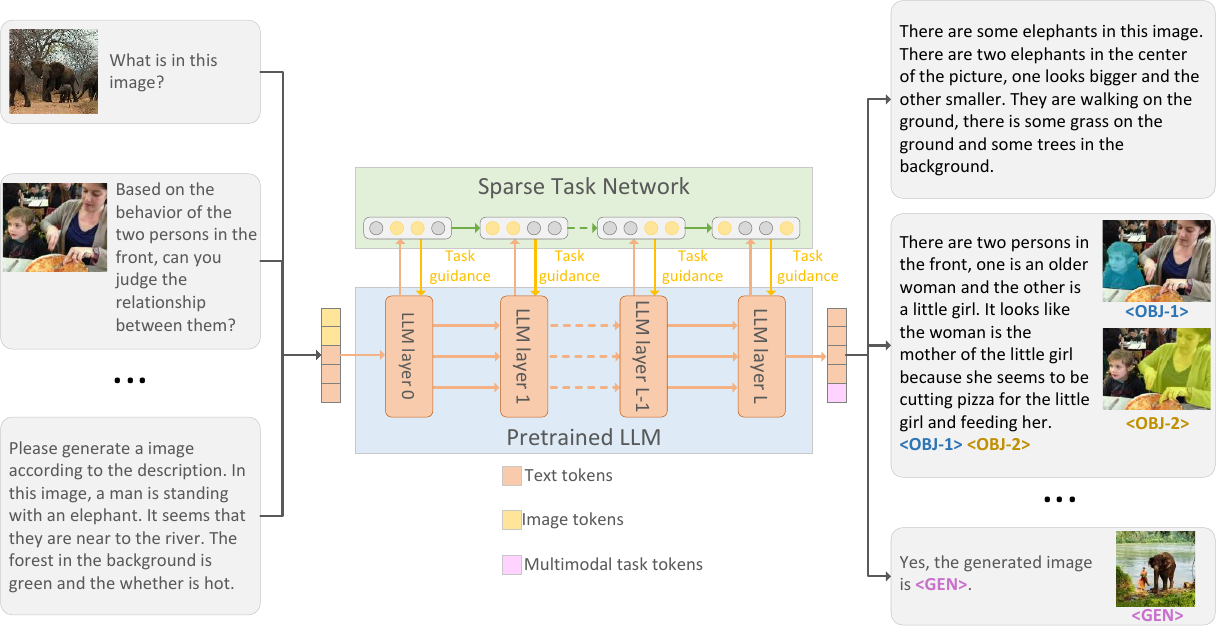}
\caption{The overview of our proposed multitask and multimodal tuning framework. All inputs and outputs are token-based, encompassing both texts and images. The model generates specific tokens for different tasks, such as $<$OBJ$>$ for segmentation and $<$GEN$>$ for text-to-image generation. During tuning, a new sparse task network is introduced to emulate SDR and provide task guidance for pretrained LLMs. The entire LLM remains frozen, with only the newly introduced parameters being tunable.}
\label{fig:first}
\end{figure*}

In revisiting the human cognitive process, which serves as the inspiration for artificial intelligence, we observe that humans inherently excel at multitask learning and effortlessly adapt to new tasks. A key factor underpinning this ability is the principle of Sparse Distributed Representation (SDR), or the Sparse Coding Hypothesis (SCH), which posits that information is represented in a way where only a small fraction of neurons are active at any given time~\cite{ahmad2016neurons,spanne2015questioning}. By activating only the necessary subset of neurons, SDR reduces energy consumption and enhances the brain’s capacity to form unique, robust representations of complex and diverse inputs. Motivated by these findings in neuroscience, we propose a unified framework for multitask and multimodal learning in large language models (LLMs), as illustrated in Figure~\ref{fig:first}. Our framework introduces two key innovations:

\textbf{Unified Tokenization for Multimodal Multitask Learning.} We formulate all tasks, as well as their multimodal data inputs and outputs, into a unified token-based representation. Unlike previous approaches that rely on cross-attention mechanisms to facilitate multimodal interactions~\cite{ouyang2023slvit}, our framework directly feeds the tokenized inputs into the model, leveraging the pretrained model’s self-attention mechanism. This allows the model to compute relationships across and within modalities in a holistic manner, enhancing its ability to understand inter- and intra-modal dependencies. By unifying tasks and modalities in this manner, we train all multimodal tasks simultaneously, enabling the model to capture shared patterns across different input data. This approach not only simplifies the tuning process but also mitigates task-specific biases, as the model learns to generalize from diverse input-output mappings. Consequently, our method allows pretrained models to handle multiple tasks with a single causal language modeling objective, significantly reducing computational overhead compared to prior methods. Furthermore, by avoiding additional cross-attention mechanisms and complex decoders, our framework ensures simplicity and computational efficiency while maintaining strong performance across tasks.

\textbf{Neural Tuning Strategy.} To further improve the efficiency and adaptability of multimodal multitask learning, we introduce neural tuning, a novel tuning strategy inspired by SDR. In this strategy, only a subset of neurons is activated for each task, mimicking the sparse activation patterns observed in the human brain. Neurons activated by different tasks consist of two components: shared neurons, which capture common features across tasks, and task-specific neurons, which encode unique characteristics for individual tasks. 
This approach aligns with the intrinsic strengths of SDR, enabling efficient multitask learning by minimizing task interference and reusing shared features while reserving task-specific pathways. By emulating this biologically inspired mechanism, our framework enhances scalability and adaptability, making it particularly suitable for handling the complexity of modern multimodal and multitask challenges in LLMs.

In summary, our framework seamlessly integrates multimodal and multitask learning into a unified architecture inspired by human cognitive principles, enabling efficient task handling with reduced computational requirements. Despite these advancements, there is currently a lack of datasets specifically designed for multimodal multitask learning, particularly for challenging tasks like reasoning segmentation, where multiple objects can only be segmented through complex reasoning that combines both image and textual information. To address this gap and foster progress in this domain, we introduce a novel dataset, MMUD (Multimodal Understanding Dataset). MMUD is constructed using GPT-4 to generate initial annotations, which are subsequently refined and verified through human annotation. The dataset comprises over 36,000 samples, each containing an image paired with a detailed content caption, a complex reasoning question-answer pair, and referring segmentation masks that align with object words in intricate descriptions. MMUD is explicitly designed to support diverse multimodal tasks that demand advanced reasoning and multimodal understanding. A comprehensive description of the dataset and its construction process is provided in Section~\ref{sec:dataset}. To demonstrate the capabilities of our framework, we fine-tune pretrained LLMs on MMUD for four distinct tasks: vanilla referring segmentation, reasoning segmentation, image captioning, and text-to-image generation. Figure~\ref{fig:first} illustrates the pipeline of our proposed method, incorporating neural tuning. Experimental results indicate that our approach achieves state-of-the-art performance across these tasks, highlighting the effectiveness and generalizability of the framework.

Our contributions are threefold:
\begin{itemize}
\item We propose a novel framework that unifies diverse multimodal tasks using a concise all-in-token methodology. This approach simplifies the integration of new tasks by requiring only the introduction of task-specific tokens, significantly enhancing the flexibility and scalability of large multimodal models.
\item We introduce neural tuning, a sparse task-tuning strategy inspired by Sparse Distributed Representation (SDR). This approach adaptively activates specific subsets of neurons for different tasks, enabling efficient multitask management while enhancing precision and adaptability across tasks.
\item We present a new multimodal benchmark, MMUD, which includes meticulously annotated samples designed for multiple tasks, such as reasoning segmentation, image captioning, and text-to-image generation. By fine-tuning models on MMUD using our proposed framework, we demonstrate superior multitask and multimodal processing capabilities, setting a new state-of-the-art in performance.
\end{itemize}

The remainder of this paper is organized as follows. In Section 2, we review related works, focusing on advancements in multimodal learning, tuning strategies, and their applications in large language models. Section 3 presents our proposed unified framework, detailing its innovative all-in-token methodology and the neural tuning strategy for efficient multitask and multimodal learning. In Section 4, we introduce the MMUD dataset, describing its construction, annotation process, and utility for evaluating multitask and multimodal learning. Section 5 illustrates the experimental results and provides a comprehensive analysis of the proposed framework's performance across various tasks. Finally, Section 6 concludes the paper by summarizing our contributions and discussing potential directions for future research.

\section{Related Works}
In this section, we analysis some recent or foundational works related to our method, including multimodal tuning, referring segmentation, and text-to-image synthesis.

\subsection{Multimodal Tuning for Large Models}

LLMs exhibit exceptional versatility across domains, but direct multimodal training is often limited by substantial hardware and data demands. To mitigate this, multimodal tuning strategies leverage pre-trained LLMs without altering their core parameters. For example, BLIP-2~\cite{li2023blip} employs a Q-Former to align image and text embeddings, and FROMAGe~\cite{koh2023grounding} uses linear projections to bridge modalities. LVLMs like LLaVA~\cite{liu2023improved} and Qwen-VL~\cite{bai2023qwen} enhance reasoning via cross-attention mechanisms, unifying input representations across modalities. However, these methods mainly focus on modality interaction at the input level. VisionLLM v2~\cite{wu2024visionllm} advances this by enabling unified processing across tasks, though it lacks task-specific fine-tuning strategies. In contrast, our framework introduces an “all-in-token” paradigm and a novel neural tuning approach tailored for multitask learning. This removes the need for explicit cross-modal mappings, simplifying the architecture while capturing both inter- and intra-modal relations through a unified self-attention mechanism. Unlike VisionLLM v2~\cite{wu2024visionllm}, our method emphasizes multitask efficiency, enabling parameter-efficient tuning with strong performance across segmentation and reasoning tasks. While recent all-in-token approaches propose unified multimodal tokens, our key innovation lies in the neural tuning mechanism specifically designed for multitask adaptation—achieving scalability, efficiency, and improved accuracy across diverse tasks.

\subsection{Referring Segmentation and Reasoning Segmentation}

Referring segmentation, a key multimodal task, involves segmenting image regions based on textual instructions, testing a model’s ability to align fine-grained visual details with language. Earlier methods, such as LAVT~\cite{yang2022lavt} and SLViT~\cite{ouyang2023slvit}, paired text encoders with U-Net-like~\cite{ronneberger2015u} vision backbones to generate masks. However, with the rise of large models, simple referring instructions pose limited challenges. To address this, complex reasoning segmentation has emerged—requiring models to answer detailed image-related questions and provide corresponding segmentations. For instance, LISA~\cite{lai2023lisa} introduces segmentation-specific tokens, while PixelLM~\cite{ren2023pixellm} uses a custom codebook for multi-instance segmentation. In our framework, reasoning segmentation serves as a core benchmark for evaluating multimodal reasoning and understanding.

\subsection{Text-to-Image Synthesis}
For text-to-image synthesis, there are generally two approaches highly related to our work: vector quantized generative adversarial network (VQGAN) related methods~\cite{esser2021taming} and diffusion-based methods~\cite{ho2020denoising,song2020score}. VQGAN aims to map images into a discrete latent space while diffusion models simulates a diffusion process, where data is gradually transformed from a simple prior distribution (like Gaussian noise) to the complex target distribution. State-of-the-art text-to-image models such as DALL-E~\cite{ramesh2021zeroshot} leverage advanced techniques in VQGAN and diffusion models to generate high-quality, diverse images. In our work, we primarily employ VQGAN for synthesizing images, but we also explore the potential of combining our approach with diffusion networks for image generation.

\begin{figure*}[t]
\centering
\begin{subfigure}{0.42\linewidth}
    \centering
    \includegraphics[width=1.0\textwidth]{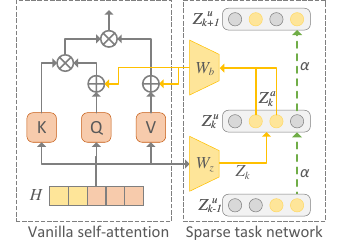}
    \caption{Detailed structure of our proposed neural tuning in self-attention.}
    \label{fig:detail}
\end{subfigure}
\begin{subfigure}{0.57\linewidth}
    \centering
    \includegraphics[width=1.0\textwidth]{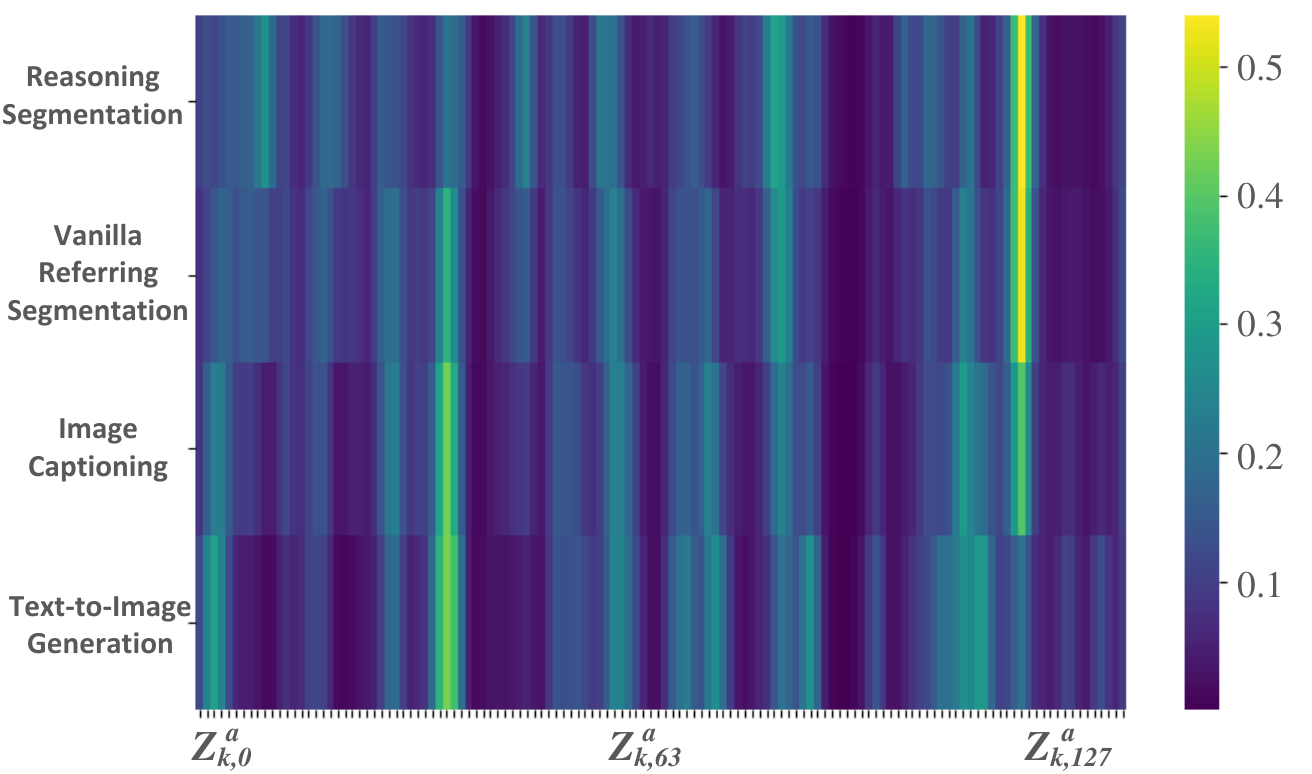}
    \caption{The visualization of the activated neurons in sparse vector.}
    \label{fig:detail2}
\end{subfigure}
\caption{The detail of proposed neural tuning and the illustration of activated neurons in sparse vector. Figure (a) illustrates the integration of the sparse task network designed to generate task guidance for pretrained LLMs. At the core of the sparse task network is the sparse vector $Z_{k}$, with only a percentage of neurons activated to perform specific tasks. Sparse vectors across different layers are interconnected through an EMA updating mechanism. Figure (b) visualized the sparse vector $Z_{k}^{a}$ for different tasks ($D_{z}$ is set to 128 in the example).}
\end{figure*}

\section{Method}
In this section, we first describe the detailed design of our unified framework with neural tuning, and then outline the specifics of multitask training.

\subsection{All-in-token Multimodal Paradigm}~\label{sec:framework}
The overall pipeline of our proposed all-in-token framework is illustrated in Figure\ref{fig:first}. To align with humans' ability to handle multiple tasks and modalities concurrently, inputs from different modalities are tokenized and processed in parallel. Inspired by the Sparse Distributed Representation (SDR) or Sparse Coding Hypothesis (SCH) of the human brain—where only a subset of neurons is activated for a specific task—we also introduce a novel sparse task network into pretrained models (further discussed in Section~\ref{sec:tuning}). This paradigm enables the model to process both image and text inputs and generate task-specific outputs, such as multi-instance referring segmentation ($<$OBJ$>$) and image generation ($<$GEN$>$).

Specifically, for a multimodal input consisting of images and sentences, we first embed the text into $I_{txt}=\{I_{txt}^{l}\}_{l=1}^{L_t}$, where $L_t$ denotes the length of the text. For image inputs, we adapt to the pretrained image encoders by first dividing the image into patches, with each patch representing a small region of the image. We then use a frozen pretrained vision encoder (such as CLIP~\cite{radford2021learning}) to extract features from these patches, resulting in $I_{img}=\{I_{img}^{l}\}_{l=1}^{L_i}$, where $L_i$, where $L_i$ denotes the number of image patches.

We then use a pretrained vision encoder to extract features from the image, resulting in $I_{img}=\{I_{img}^{l}\}_{l=1}^{L_i}$, where $L_i$ represents the number of image patches. To integrate the multimodal input, we concatenate the text embeddings and image features (when visual input is required for the task) to form the final input for the pretrained large language model (LLM): $I = [I_{img}; I_{txt}]$. In this scheme, the textual and visual modalities interact through the LLM's self-attention mechanism, where the output $O$ is computed as $O=SoftMax(\frac{QK^T}{\sqrt{d_k}})V$ with the Query($Q$), Key($K$), and Value($V$) derived from the concatenated multimodal input $I$. 

Unlike prior works that use cross-attention for modality interaction~\cite{ouyang2023slvit, yang2022lavt}, where $Q$, $K$, and $V$ are derived from different modalities, our all-in-token approach offers several advantages. It not only simplifies the architecture by avoiding explicit cross-modal mappings, but also enables the model to compute both inter- and intra-modal relationships. This dual interaction within the self-attention framework enhances the model’s ability to effectively understand and fuse multimodal information. Furthermore, when incorporating additional modalities (such as audio), this approach eliminates the need for designing complex cross-attention schemas and allows for seamless extension to new modalities by simply concatenating the corresponding modality tokens.

In the output stage, we introduce new task-specific tokens alongside the original textual tokens to handle multimodal tasks. For instance, we incorporate $<$OBJ$>$ tokens for segmentation tasks and $<$GLB$>$ tokens for text-to-image synthesis. These task-specific tokens are then passed to their respective decoders, enabling the model to generate appropriate outputs for each task. This approach unifies the input and output formats across all tasks into an all-in-token scheme, simplifying the integration of additional tasks and modalities. Consequently, the system becomes more flexible and scalable, allowing for seamless extension to new multimodal tasks without significant architectural modifications.

\subsection{Neural Tuning for Large Models}~\label{sec:tuning}
To fine-tune the pretrained LLMs, we introduce a sparse task network behaving like SDR. It is parallel to pretrained LLMs but gets linked in each layer for task guidance. For each tuning layer, it maintains a learnable vector named the \textit{Sparse Vector} and the details are shown in Figure~\ref{fig:detail}. In the $k$-th layer of the LLM before self-attention, we first project the hidden embeddings $H$ into a subspace to obtain the sparse vector:
\begin{equation}
    Z_k = W_z H \in R^{(L_t+L_i)\times D_z},
\end{equation}
where $H\in \mathbb{R}^{(L_t+L_i)\times D_h}$, $W_z\in \mathbb{R}^{D_z \times D_h}$ is a learnable matrix, $D_h$ is the embedding size of the pretrained LLM, and $D_z$ is the dimension of the sparse vector. To ensure the sparse vector is transmitted between layers, $Z_k$ is updated using an Exponential Moving Average (EMA) mechanism with a hyperparameter $\alpha$:
\begin{equation}
\label{equ:update}
    Z_k^u = \alpha Z_{k-1}^{u} + (1-\alpha)Z_k,
\end{equation}
where $Z_{k-1}^u$ is the sparse vector from the last layer. This approach allows the sparse vector to not only receive information from the previous layer but also be aware of the current layer's information. To enable sparse distributed representation like human brain during model flow, only a subset of the sparse vector's nodes are activated for various tasks. We first randomly sample an activation rate $r$ from a normal distribution $p(r) = N(r; \beta, (0.1\beta)^2)$, where $\beta\in (0, 1)$ is a predefined hyperparameter\footnote{More activation patterns are discussed in Section~\ref{app:active}}. Then, we activate corresponding neurons as follows:
\begin{equation}
    Z_{k, j}^a = 
    \left\{
    \begin{matrix}
    \begin{aligned}
      & Z_{k, j}^u \quad & Z_{k, j}^u >= Z_{k, r}^u \\  
      & 0          \quad & Z_{k, j}^u <  Z_{k, r}^u 
    \end{aligned}
    \end{matrix}
    \right.
\end{equation}
where $Z_{k, r}^u$ is the largest top $r$ values and $j\in [1, D_z]$. $Z_{k}^a$ represents the activated neurons for a specific task. Next, to allow the pretrained LLM to leverage task-specific guidance, we use a linear transformation to project the activated sparse vector back to the LLM's hidden space:
\begin{equation}
    Z_k^b = W_b Z_k^a \in \mathbb{R}^{(L_t+L_i)\times (2D_h)},
\end{equation}
where $W_b$ is a learnable parameter. For the self-attention mechanism of the LLM, we split $Z_k^b$ into two parts and use the residual for query ($Q$) and value ($V$) computation,
\begin{equation}
\label{equ:infuse}
    \begin{aligned}
    Q' &= W_q H + Z_k^b[:, 0:D_h], \\
    V' &= W_v H + Z_k^b[:, D_h:2D_h], \\
    K  &= W_k H. 
    \end{aligned}
\end{equation}
$Q'$, $K$, and $V'$ are then employed for the vanilla self-attention in pretrained LLMs. Overall, the pretrained LLM is responsible for the main inference, while the sparse task network handles task-specific execution, just like different parts in human brains.

Building on the described activation mechanism, neurons activated by different tasks consist of two components: shared neurons, which capture common features across tasks, and task-specific neurons, which encode the unique characteristics of individual tasks. Interestingly, our experiments reveal that tasks with higher relatedness tend to activate a greater number of shared neurons $Z_k^a$. For example, tasks like vanilla referring segmentation and reasoning segmentation exhibit significant overlap in neuron activation patterns, as they share common underlying features, such as object detection and spatial reasoning. Figure~\ref{fig:detail2} provides a detailed visualization of these activation patterns. However, as we could draw from the visualization, some neurons are rarely activated across different tasks. This phenomenon can be attributed to two factors. Firstly, certain neurons are highly specialized for tasks that are less emphasized or represented in current experimental setups, resulting in fewer activations. Secondly, sparse activation ensures that only the most relevant neurons are utilized for a given task. This approach not only enhances computational efficiency but also prevents over-reliance on any single component, enabling the model to maintain adaptability across diverse tasks. 

This visualization not only supports the design rationale of our sparse task network but also offers insights into how the model dynamically allocates resources to shared and task-specific components. Such behavior aligns with the human brain’s efficiency in reusing common cognitive processes while adapting to task-specific demands.

% This dynamic allocation of shared and task-specific neurons enables the framework to maintain scalability and flexibility while preserving task performance. Moreover, the visualization offers a clear explanation of task relationships, providing a foundation for further optimization of multitask learning systems.

\subsection{Multitask Training}
\begin{figure*}[t]
\centering
\includegraphics[width=1.0\textwidth]{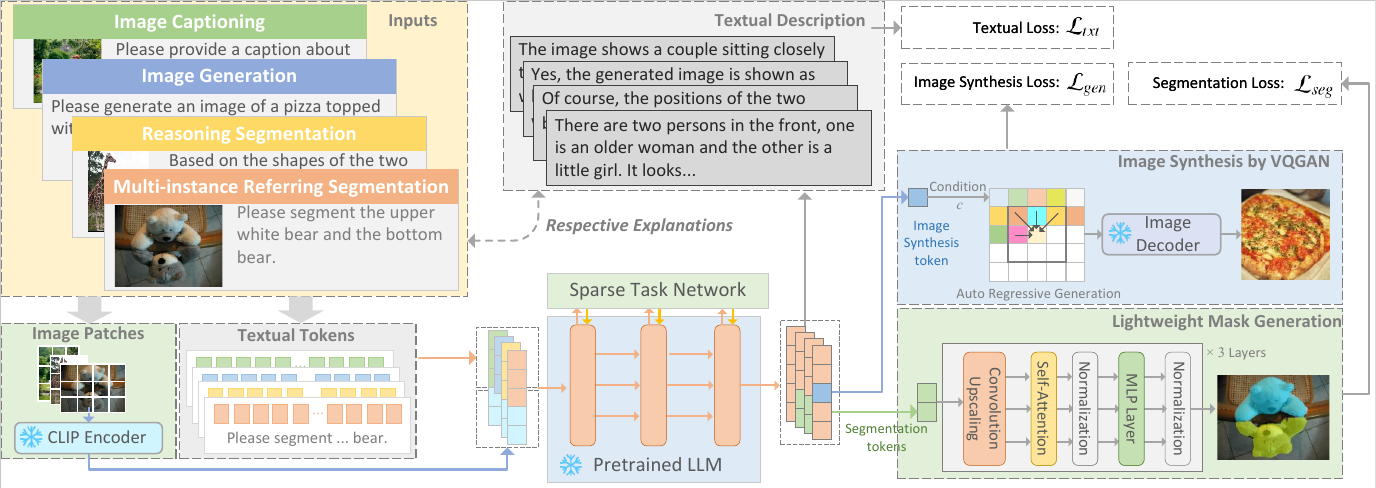}
\caption{The pipeline of our proposed framework for multitask training. After processing multimodal tokens, the LLM generates corresponding task tokens along with respective explanations. For segmentation-related tasks, a lightweight three-layer decoder is used for mask generation, while for image synthesis tasks, VQGAN is employed to generate images in an autoregressive manner.}
\label{fig:pipeline}
\end{figure*}

The whole pipeline for our proposed multitask training is shown in Figure~\ref{fig:pipeline}. As different tasks are unified into an all-in-one token manner, the tuning procedure can be conducted using a simple causal language modeling approach. Cross-entropy loss is employed as the loss function ($\mathcal{L}_{txt}$) for next-token prediction:
\begin{equation}
\mathcal{L}_{txt}=-\sum_t \log \hat{P}(x_t|x_i; i<t), %_{t=1}^{L_t}
\end{equation}
where $\hat{P}(x_t|x_i; i<t)$ is the predicted probability for the token $x_t$ based on the context of all previous tokens.

For segmentation-related tasks, while it is feasible to use the same architecture as the image-synthesis task, we prioritize efficiency by proposing a separate lightweight decoder for segmentation mask generation. Specifically, the embeddings corresponding to object tokens, $<$OBJ$>$, denoted as $H_{seg}$, are extracted and fed into the lightweight decoder:
\begin{equation}
\hat{y}=Decoder_{seg}(H_{seg}, W_{seg}),
\end{equation}
where $W_{seg}$ represents the learnable parameters in the mask decoder. Different from previous methods \cite{lai2023lisa}, which employ pretrained SAM~\cite{kirillov2023segment} for predicting masks, our designed decoder is much lighter. The mask decoder consists of three convolutional layers for upscaling, each followed by three self-attention layers for mask decoding. Since the models can generate multiple segmentation tokens at a time, we can easily perform multi-instance segmentation. Following previous works~\cite{lai2023lisa,ren2023pixellm}, DICE loss is employed in our framework to guide the segmentation tasks:
\begin{equation}
\begin{aligned}
\mathcal{L}_{seg} = \mathcal{L}_{DICE} = \frac{1}{N}\sum_{i}(1- \frac{2|\hat{y_i}\cap y_{gt,i}|}{|\hat{y_i}|+ |y_{gt,i}|})
\end{aligned}
\end{equation}
where $N$ is the number of samples, $\hat{y}$ is the predicted mask and $y_{gt}$ is the ground truth. 

In text-to-image synthesis tasks, we employ pretrained VQGAN generators~\cite{esser2021taming,wang2022clip} to synthesize images. To generate a sequence of indices for VQGAN to produce images, we train a conditional transformer to predict the indices in an autoregressive manner:
\begin{equation}
p(\mathbf{s} |c) = \prod_i p(s_i|s_{<i}, c),
\end{equation}
where $\mathbf{s}$ is the sequence of indices for VQGAN to generate images and $c$ is the condition that controls the contents of the images. To simplify the tuning process, we first pretrain the conditional transformer on MS-COCO~\cite{lin2014microsoft} and employ the image embeddings from the pretrained CLIP model~\cite{radford2021learning} as the condition. After training, we can generate images via the CLIP embeddings. Therefore, in neural tuning, we simply align the embeddings of the synthesis tokens ($H_{gen}$) to the CLIP embeddings ($H_{clip}$). The mean squared error is employed to perform the alignment:
\begin{equation}
    \mathcal{L}_{gen} = \mathcal{L}_{MSE} = \frac{1}{N}\sum_{i}(W_{gen}H_{gen}-H_{clip})^2,
\end{equation}
where $W_{gen}$ is a learnable transformation to project token embeddings into the CLIP feature space.

The pretrained LLMs are then tuned with all tasks involved. The overall training loss is represented as:
\begin{equation}
\mathcal{L} = \mathcal{L}_{txt} + \lambda_{seg}\mathcal{L}_{seg} + \lambda_{gen}\mathcal{L}_{gen},
\end{equation}
where $\lambda_{seg}$ and $\lambda_{gen}$ are coefficients to balance the numerical scales of different losses.
Although segmentation and generation tasks are the primary illustration for multitask scalability in our current approach, the method can be easily extended to additional tasks by introducing new corresponding task-specific tokens.

\begin{figure*}[t]
\centering
\includegraphics[width=1.0\textwidth]{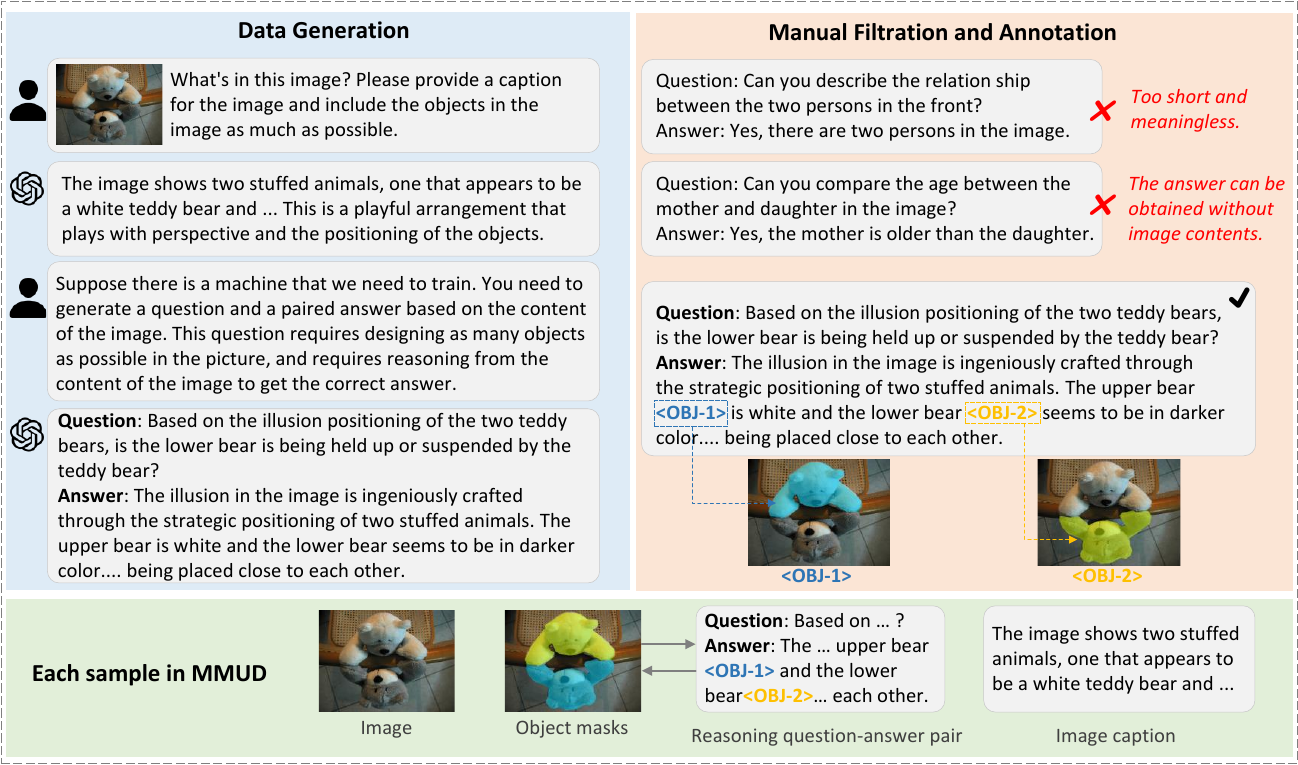}
\caption{Pipeline for MMUD Dataset Generation. First, GPT-4v is used to generate captions describing the image contents and reasoning question-answer pairs. Then to ensure dataset quality, meaningless cases are filtered out. Finally, $<$OBJ$>$ tokens are manually appended to objects in the answers to help large models better understand the relationships between images and text.}
\label{fig:dataset}
\end{figure*}

\section{MMUD Benchmark}~\label{sec:dataset}
To enable LLMs with multitask and multimodal processing capabilities, a high-quality dataset is essential. However, there are few datasets that provide multitask annotations specifically designed for large models. To address this, we have constructed the MMUD dataset. MMUD dataset contains over 36,000 samples. We have divided the dataset into training, validation, and test subsets, comprising 33,682, 1,400, and 1,400 samples, respectively. Each sample in the dataset is annotated with referring segmentation masks, complex reasoning question-answer pairs, and image captions. In this work, we focus on four tasks with MMUD: vanilla referring segmentation, reasoning segmentation, image captioning and text-to-image generation.

We constructed our dataset using open-source datasets: RefCOCO~\cite{yu2016modeling}, RefCOCOg \cite{mao2016generation}, and RefClef~\cite{nagaraja2016modeling,mao2016generation}. Since these datasets only provide referring segmentation masks, we augmented them with annotations for other tasks. The dataset generation pipeline is illustrated in Figure~\ref{fig:dataset}. Initially, we employed GPT-4v to generate image captions describing the contents. Subsequently, GPT-4v was used to generate complex questions along with answers based on the image contents. We then performed manual filtration to ensure that the generated contents were meaningful and suitable for multimodal understanding. We filtered out samples in cases where: 1) the generated contents were meaningless; 2) the length of the generated contents was either too long or too short; 3) the answers could be directly inferred from the questions without reference to the images. To enhance the capability of complex reasoning, we manually inserted the $<$OBJ-i$>$ token after each object in the answer, where $i$ represents the $i$-th object in the image. For example, when generating the question \textit{Can you judge the relationship between the two people in the front based on what happened in the picture?" with the answer} with answer \textit{In this image, one of the women in the front is cutting a pizza, and a little girl is next to her...}, we manually inserted the $<$OBJ$>$ token after the expression of each object: \textit{In this image, one of the women in the front $<$OBJ-1$>$ is cutting a pizza $<$OBJ-2$>$, and a little girl $<$OBJ-3$>$ is next to her...}. Following the structure of the RefCOCO datasets, each sample image contains multiple meaningful objects with corresponding masks. Each $<$OBJ-i$>$ token not only corresponds to an object but also to the corresponding segmentation mask, which could help models better capture the relationship between texts and images. The dataset is publicly available at https://github.com/kiva12138/NeuralTuning.

\section{Experiments and Analysis}
% In this section, we illustrate the details of implementations of neural tuning and shows experimental results of our method compared with previous methods.

\subsection{Experimental Settings}
In our method, we employ the pretrained LLaMA2-13B and LLaMA2-7B as the textual foundation models. For image feature extraction, we utilize CLIP-ViT-L/14. The images are resized to 224 pixels as inputs. All parameters in LLaMA and CLIP are kept frozen, and only the newly introduced tokens, sparse task network, and task decoders are trainable (2.9\% in total). The efficient parameter $\alpha$ in sparse vector updating is set to 0.9, while the neuron activation rate $\beta$ is set to 0.4. We set the dimension of the sparse vector $Z$ to 128. During optimization, to balance the numerical scales in the loss functions, we set $\lambda_{seg}$ to 1.0 and $\lambda_{gen}$ to 10.0. We train the models for 10 epochs with a batch size of 12 and a cosine learning rate decay scheduler. The tuning process takes about 36 hours on four RTX 4090 GPUs or three NVIDIA A100 GPUs.

\begin{table}[t]
\centering
\caption{The summary of our our evaluated tasks, datasets, metrics, and corresponding results.}
\label{result_summary}
\resizebox{0.5\linewidth}{!}{
\begin{tabular}{c|c|c|c}
\toprule
Tasks & Datasets & Main Metric & Main Results \\
\midrule
\multirow{3}{*}{\makecell{Referring \\ Segmentation}} & RefCOCO & \multirow{5}{*}{\makecell{oIoU}} & 74.2 \\
 & RefCOCO+ &  & 66.3 \\
 & RefCOCOg &  & 70.3 \\
 % \midrule
\cmidrule{1-2} \cmidrule{4-4}
\makecell{Reasoning \\ Segmentation} & \multirow{6}{*}{MMUD} &  & 64.2 \\
\cmidrule{1-1} \cmidrule{3-4}
\makecell{Image\\ Captioning} &  & BLEU-4 & 43.7 \\
\cmidrule{1-1} \cmidrule{3-4}
\makecell{Image\\ Synthesis} &  & FID & 12.7    \\
\midrule
\multirow{2}{*}{\makecell{VQA}} & VQA v2 & \multirow{2}{*}{\makecell{VQA}} & 72.1 \\
 & TextVQA & & 57.6 \\
\bottomrule
\end{tabular}}
\end{table}

\begin{table*}[h]
\centering
\caption{The results our neural tuning (NT in table) and previous methods on MMUD and three public datasets for reasoning segmentation and referring segmentation.}
\label{result_seg}

% \resizebox{1.0\linewidth}{!}{
\begin{tabular}{c}

\begin{subtable}[h]{0.95\textwidth}
    \centering
    \caption{The results of reasoning segmentation on MMUD dataset. As MMUD is a new dataset, we re-implement the methods based on their open-source codes.}
    \label{result_seg_mmud}
    \begin{tabular}{cccccc}
        \toprule
        \multirow{2}{*}{Method} & \multirow{2}{*}{w/ LLM} & \multicolumn{2}{c}{Valid} & \multicolumn{2}{c}{Test} \\
         &  & mIoU & oIoU & mIoU & oIoU \\
        \midrule
        LAVT &  & 21.2 & 20.2 & 23.3 & 23.1 \\
        LISA-7B & $\checkmark$ & 60.1 & 59.9 & 61.3 & 61.8 \\
        PixelLM-7B & $\checkmark$ & 61.1 & 60.7 & \textbf{63.2} & 62.6 \\
        NT-7B(Ours) & $\checkmark$ & \textbf{62.2} & \textbf{61.6} & 63.1 & \textbf{62.8} \\
        \midrule
        LISA-13B & $\checkmark$ & 62.0 & 61.2 & 63.1 & 62.7 \\
        PixelLM-13B & $\checkmark$ & \textbf{63.4} & 62.8 & 64.4 & 64.0 \\
        NT-13B(Ours) & $\checkmark$ & \textbf{63.4} & \textbf{63.0} & \textbf{64.9} & \textbf{64.2} \\
        \bottomrule
    \end{tabular}
\end{subtable} \\

\begin{subtable}[h]{1.0\textwidth}
    \centering
    \caption{The results of vanilla referring segmentation on RefCOCO, RefCOCO+, and RefCOCOg are presented. The metric used in the table is oIoU.}
    \label{result_seg_public}
    \resizebox{1.0\textwidth}{!}{
    \begin{tabular}{ccccccccccc}
        \toprule
        \multirow{2}{*}{Method} & \multirow{2}{*}{w/ LLM}& \multirow{2}{*}{TFLOPs} & \multicolumn{3}{c}{RefCOCO} & \multicolumn{3}{c}{RefCOCO+} & \multicolumn{2}{c}{RefCOCOg(U)} \\
        & &  & Val & TestA & TestB & Valid & TestA & TestB & Valid & Test \\
        \midrule
        LAVT & & & 72.7 & 75.8 & 68.8 &62.1 & 68.4 & 55.1 & 61.2 & 62.1 \\
        LISA-7B & $\checkmark$ & 7.16 & 74.0 & 76.3 & 70.4 & 62.5 & 66.3 & 56.0 & 67.0 & 69.1 \\
        PixelLM-7B & $\checkmark$ & 3.57 & 73.0 & 76.5 & \textbf{68.2} & \textbf{66.3} & \textbf{71.7} & \textbf{58.3} & 69.3 & \textbf{70.5} \\
        NT-7B(Ours) & $\checkmark$ & \textbf{2.47} & \textbf{74.2} & \textbf{76.7}& 68.0 & \textbf{66.3} & 71.2 & 58.1 & \textbf{70.3} & 70.2 \\
        \bottomrule 
    \end{tabular}}
\end{subtable}

\end{tabular}
% }
\end{table*}

\subsection{Results on MMUD}
Table~\ref{result_summary} summarizes the evaluated tasks, datasets, main metric and corresponding results for our proposed framework. In detail, Table~\ref{result_seg}, Table~\ref{result_cap}, and Table~\ref{result_vqa} presents the comparison with previous methods. For vanilla referring segmentation and reasoning segmentation, we mainly compare our method with LAVT~\cite{yang2022lavt}, LISA~\cite{lai2023lisa}, and PixelLM~\cite{ren2023pixellm}. We employ the mean intersection over union (mIoU) and overall IoU as the metrics. LAVT aims to fuse the BERT features into vision backbones, achieving great results in vanilla segmentation. However, when it comes to complex reasoning scenarios, LAVT fails to converge while other LLM-based methods yield good results, demonstrating the powerful reasoning capabilities of large models. Compared with LISA and PixelLM, which employ pretrained LLaVA~\cite{liu2023improved} and focus on complex reasoning segmentation, neural tuning can achieve better results, demonstrating its effectiveness. Regarding inference speed, our tuned 7B model can complete the reasoning segmentation process within 80ms per text-image pair on an RTX 4090 GPU, while the 13B model takes 110ms on an RTX A6000 GPU. This demonstrates the efficiency of our proposed neural tuning. We believe this efficiency results from the linear sparse vector updating and task guidance. 

Furthermore, to further evaluate the effectiveness of our proposed method, we also present the performance on the original test set of RefCOCO, RefCOCO+, and RefCOCOg datasets~\footnote{For fair comparison, we re-split the train, validation, and test set of MMUD according to the original datasets.}. As shown in Table~\ref{result_seg_public}, our method achieves state-of-the-art performance on the public datasets but with lower computational burden, revealing the effectiveness and efficiency of our proposed multitask neural tuning for segmentation tasks.

\begin{table*}[t]
\centering
\caption{The results our neural tuning and previous methods on MMUD for image captioning task and text-to-image generation task.}
\label{result_cap}

\resizebox{1.0\linewidth}{!}{
\begin{tabular}{cccc|cccc}
\toprule
\multicolumn{4}{c|}{Image Captioning} & \multicolumn{4}{c}{Text-to-image Synthesis} \\
\midrule
Method & BLEU-4 & METEOR & CIDEr & Method & FID($\downarrow$) & KID($\downarrow$) & IS \\
\midrule
BLIP2-6.7B & 41.9 & 34.7 & 133.0 & GLIGEN & 12.9 & \textbf{12.5} & 31.1 \\
ExpansionNetV2 & 41.1 & 34.0 & 132.7 & U-ViT-S/2 & 13.7 & 15.9 & 29.8 \\
mPLUG & 43.0 & 34.1 & \textbf{134.0} & Parti & \textbf{11.0} & 13.6 & 30.6 \\
NT-7B(Ours) & \textbf{43.7} & \textbf{35.5} & 133.2 & NT-7B(Ours) & 12.7 & 15.2 & \textbf{31.4} \\
\bottomrule 
\end{tabular} }
\end{table*}

\begin{table}[t]
\centering
\caption{The zero-shot performance of our proposed framework with previous methods on VQA v2 (test-dev) and TextVQA (val) datasets for visual question answering task.}
\label{result_vqa}
\resizebox{0.5\linewidth}{!}{
\begin{tabular}{ccc}
\toprule
% \multicolumn{3}{c}{VQA Performance} \\
% \midrule
Method &  VQA v2 & TextVQA \\
\midrule
BLIP-2~\cite{li2023blip} & 41.0 & 42.5 \\
InternVL-Chat~\cite{chen2024internvl} & 72.3 & 42.1  \\
InstructBLIP~\cite{instructblip} & - & 50.1  \\
SPHINX-Intern2~\cite{liu2024sphinx} & 75.5 & 58.1 \\
VisionLLM v2~\cite{wu2024visionllm} & 80.8 & 64.7 \\
NT-7B(Ours) & 72.1 & 57.6  \\
\bottomrule 
\end{tabular}}
\end{table}

\begin{table*}[t]
\centering
\caption{The ablation study on MMUD test set for reasoning segmentation. The LLaMA2-7B is employed in the ablation. $Z_k^U$ Upd. means the sparse vector updating module (Equation~\ref{equ:update}) and SDR indicates the sparse distributed representation emulation. Task V.S., Task I.C. and Task I.S. means the vanilla referring segmentation, image captioning and text-to-image synthesis tasks, respectively.}
\label{result_ablation}

\resizebox{1.0\linewidth}{!}{
\begin{tabular}{ccccc|ccc|cc}
\toprule
\multicolumn{5}{c|}{Ablations} & \multicolumn{3}{c|}{Tuning Modules} & \multicolumn{2}{c}{MMUD} \\
$Z_k^U$ Upd. & SDR & Task V.S. & Task I.C. & Task I.S. & Q & K & V & mIoU & oIoU \\
\midrule
$\checkmark$ &  &  &  &  & $\checkmark$ & & $\checkmark$ & 57.7 & 57.6 \\
 & $\checkmark$ &  &  &  & $\checkmark$ & & $\checkmark$ & 59.3 & 58.2 \\
$\checkmark$ & $\checkmark$ &  &  &  & $\checkmark$ &  & $\checkmark$ & 60.0 & 58.9 \\
\midrule
$\checkmark$ & $\checkmark$ & $\checkmark$ &  &  & $\checkmark$ &  & $\checkmark$ & 61.1 & 50.7 \\
$\checkmark$ & $\checkmark$ &  & $\checkmark$ &  & $\checkmark$ & & $\checkmark$ & 62.0 & 61.2 \\
$\checkmark$ & $\checkmark$ &  &  & $\checkmark$ & $\checkmark$ & & $\checkmark$ & 62.9 & 62.6 \\
\midrule
$\checkmark$ & $\checkmark$ & $\checkmark$ & $\checkmark$ & $\checkmark$ & $\checkmark$ & & $\checkmark$ & 63.1 & 62.8 \\
$\checkmark$ & $\checkmark$ & $\checkmark$ & $\checkmark$ & $\checkmark$ & & $\checkmark$ & $\checkmark$ & 62.0 & 61.1 \\
$\checkmark$ & $\checkmark$ & $\checkmark$ & $\checkmark$ & $\checkmark$ & $\checkmark$ & $\checkmark$ & $\checkmark$ & \textbf{63.5} & \textbf{63.0} \\
\bottomrule
\end{tabular}}
\end{table*}

For image captioning, we compare our method with BLIP-2~\cite{li2023blip}, ExpansionNetV2 \cite{hu2022expansionnet} and mPLUG~\cite{li2022mplug}. BLIP-2 proposed a Q-Former for multimodal interaction whlie mPLUG learn the relationship between modalities by cross-modality skip-connection. We employ the BLEU-4~\cite{papineni2002bleu}, CIDEr~\cite{vedantam2015cider}, and METEOR~\cite{banerjee2005meteor} metrics for evaluation. The results are shown in Table~\ref{result_cap}. As we can draw from the results, we can reach the competitive performance compared with previous state-of-the-art approaches. 

On text-to-image synthesis tasks, quantitative metrics are shown in Table~\ref{result_cap}. We employ FID score, KID Score, and inception score (IS) as the metrics. We compare our method with previous methods, including Parti~\cite{yu2022scaling}, GLIGEN~\cite{li2023gligen}, and U-ViT-S/2~\cite{bao2023all}. Compared with other methods, one of the significant advantage of our method is that neural tuning is designed for multitask tuning instead of a certain task. 

In addition, we evaluated our proposed method on visual question answering (VQA) tasks in a zero-shot manner. The experiments are conducted on VQA v2~\cite{goyal2017making} and TextVQA~\cite{singh2019towards} datasets. The results, presented in Table~\ref{result_vqa}, demonstrate the effectiveness of our approach in handling visual question answering tasks in a zero-shot manner, without task-specific fine-tuning. Notably, our method achieves competitive performance compared to state-of-the-art models such as VisionLLM v2~\cite{wu2024visionllm} and InterVL-Chat~\cite{chen2024internvl}, despite not relying on specialized tuning for individual tasks. This highlights the strong generalizability and robustness of our framework across diverse benchmarks, emphasizing its practicality for multitask and low-resource scenarios.

Despite the promising results demonstrated by our unified framework, particularly in segmentation tasks where it achieves state-of-the-art performance, we observe trade-offs in its performance on generative tasks such as image synthesis. While the framework matches or slightly underperforms compared to specialized methods~\cite{yu2022scaling,li2023gligen} in these tasks, this outcome reflects the inherent balance between generality and specialization. Unlike task-specific architectures optimized exclusively for image generation~\cite{bao2023all}, our unified framework prioritizes efficiency, flexibility, and scalability across diverse tasks. This design choice ensures robust multitask learning capabilities but may limit its ability to fully exploit the potential of generative tasks. Future work could address this trade-off by integrating advanced generative modules into the framework, enabling enhanced performance while preserving the lightweight and unified structure.

\subsection{Case Visualization}\label{app:visual}

\begin{figure*}%[t]
\centering
\includegraphics[width=1.0\textwidth]{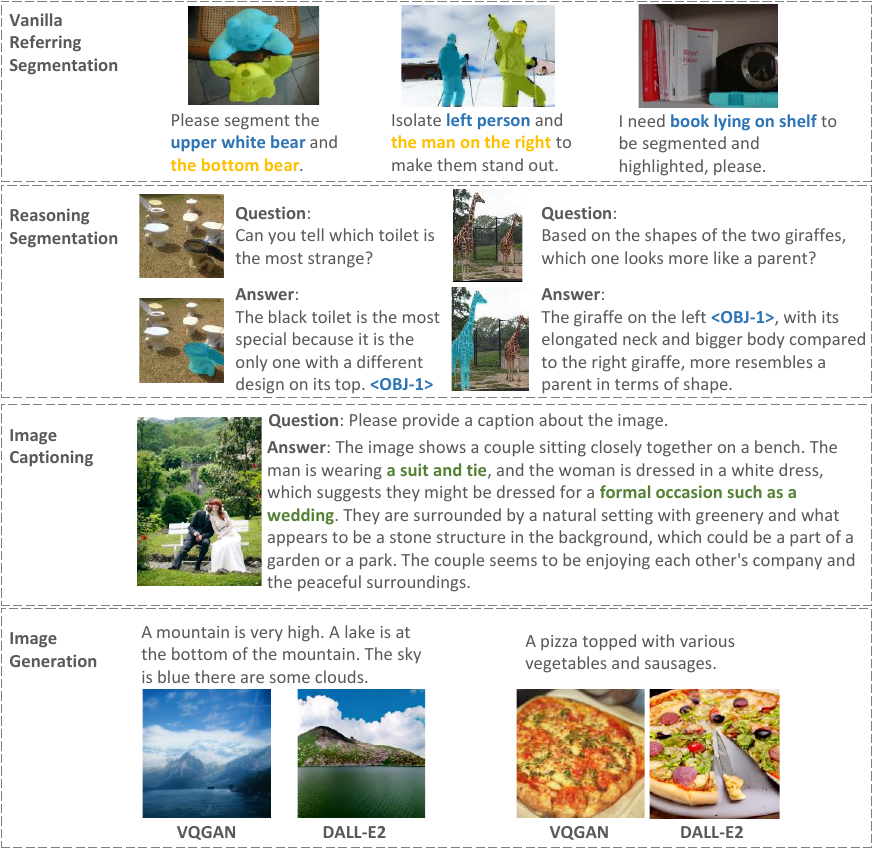}
\caption{The visualization of some cases from our model, including vanilla referring segmentation, reasoning segmentation, image captioning, and image generation. }
\label{fig:visual}
\end{figure*}

The visualizations of several cases generated by our model are shown in Figure~\ref{fig:visual}. These cases encompass a variety of tasks, including vanilla referring segmentation, reasoning segmentation, image captioning, and image generation. For image generation, we utilize two pretrained decoders: VQGAN and DALL-E2. In the case of reasoning segmentation, the model not only accurately segments the corresponding objects but also provides detailed reasoning or explanations for the results. These examples highlight the model’s ability to seamlessly handle diverse multimodal tasks, demonstrating both its precision in segmentation and its creative capabilities in image generation.

\begin{figure*}[t]
\centering
\includegraphics[width=0.9\textwidth]{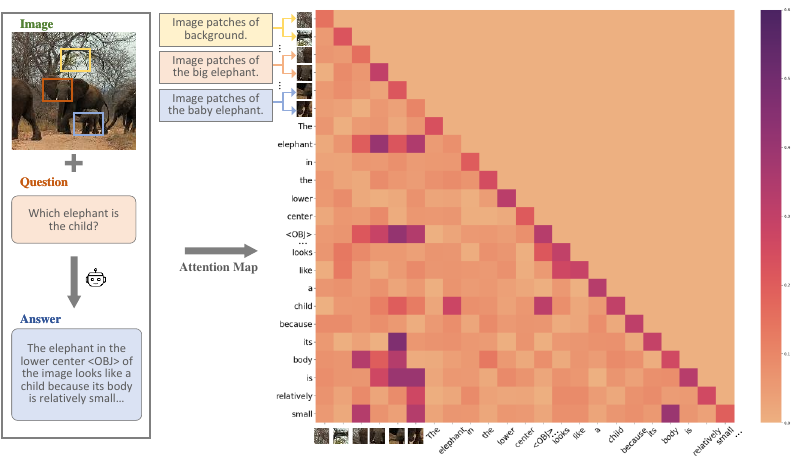}
\caption{Visualization of Attention Scores for a Reasoning Segmentation Case. For better clarity, we omit the question and answer tokens, as well as some image patches. Darker regions indicate higher attention weights. The results demonstrate that the model effectively builds connections between image details, text, and the <OBJ> task token.}
\label{fig:attention_scores}
\end{figure*}

\subsection{Effectiveness of Self-Attention in All-in-token Paradigm}
In our approach, we utilize self-attention to compute the attention scores between the various elements of the input, including both image patches and textual tokens. This differs from traditional methods that rely on cross-attention, where separate attention mechanisms are used for different modalities. The self-attention mechanism, by contrast, enables a more unified interaction between modalities, allowing the model to capture both intra-modal and inter-modal relationships simultaneously.

As shown in Figure~\ref{fig:attention_scores}, we visualize the attention scores between image patches and corresponding textual tokens for reasoning segmentation with relatively shorter sentences. The figure demonstrates that object-related words in the text (such as "cat," "car," or "tree") receive significantly higher attention weights from their corresponding image patches. This indicates that the self-attention mechanism effectively focuses on the relevant visual regions when processing the corresponding linguistic cues, facilitating a more coherent integration of multimodal information. The ability to compute these attention scores in a unified manner not only streamlines the model's architecture but also enhances its capacity to perform tasks requiring intricate cross-modal reasoning, such as referring segmentation and text-to-image synthesis.

\subsection{Ablation Study}
To prove that each module of our proposed method is effective, we conducted ablation experiments on complex reasoning segmentation tasks. The ablation of different aspects are as follows:
% All of the experiments are conducted under the same conditions. The ablation of different aspects are as follows:

\textbf{Modules in neural tuning:} The quantitative results are presented in Table~\ref{result_ablation}. We observed that removing sparse vector updating (Equation~\ref{equ:update}), which renders the tuning layers of the sparse task network independent of each other, leads to a drop in performance. This highlights the significance of interlinking different layers of tuning networks, a factor overlooked by prior tuning methods such as LoRA~\cite{hu2021lora} and (IA)$^3$~\cite{liu2022few}. Additionally, performance degrades when the SDR strategy is eliminated, indicating that all neurons are activated for all tasks. 
% We also explored the impact of varying the $\beta$ parameter, which controls the neuron activation rate in the sparse vector. Interestingly, setting $\beta$ to 0.4 resulted in improved performance when all four tasks were considered. For three tasks, the optimal performance was achieved with $\beta$ set to 0.5 or 0.6. Conversely, when one or two tasks were involved, it was more beneficial to activate a higher proportion of neurons (0.9 or 1.0). The detailed experiments are illustrated in Appendix~\ref{app:beta}. 
Finally, we conducted an ablation study on multitask learning. The results show that removing the multitask learning strategy led to a relative decrease in performance. Notably, the vanilla referring segmentation and image captioning tasks appear to contribute more to complex reasoning than the image generation tasks, which corresponds to the visualized results in Figure~\ref{fig:detail2}.

\textbf{The selection of tuning modules:} In Section~\ref{sec:tuning}, we integrated task guidance into the LLMs for query and value, following established methods~\cite{hu2021lora}. However, this approach is not the sole option. We conducted corresponding ablation studies, with the quantitative results presented in Table~\ref{result_ablation}. The results indicate that integrating task guidance into keys and values tends to result in poorer performance. While performance improves when task guidance is integrated into all query, key, and value modules, the improvement is only marginal compared to when it is integrated solely into query and value. Therefore, to strike a balance between performance and complexity, we opted to perform tuning only on query and value modules (Equation~\ref{equ:infuse}).

\textbf{Choice of image generation decoders:} Apart from VQGAN, there are alternative options for image decoding. We also utilized a pre-trained DALL-E2 decoder for generating images from the hidden embeddings. A comparison is presented in Figure~\ref{fig:visual} (Section ~\ref{app:visual}). The quality and style of the generated images are significantly influenced by the pre-trained decoders. This bias is a result of the pretraining data used for the decoders. Hence, in theory, neural tuning can employ any pre-trained image synthesis decoder. Unfortunately, due to hardware limitations, we were unable to test other pre-trained models for image generation.

\begin{table*}[t]
\centering
\caption{Ablation study of $\beta$ and $\alpha$ in sparse vector updating. The metrics reported are the oIoU scores for Reasoning Segmentation on the MMUD dataset.}
\label{ablation_beta}
\resizebox{\textwidth}{!}{
\begin{tabular}{c|cccccccccc}
\toprule
\multirow{2}{*}{Tasks involved} & \multicolumn{10}{c}{Ablation on the Selection of $\beta$} \\
 & $\beta=0.1$ & $\beta=0.2$ & $\beta=0.3$ & $\beta=0.4$ & $\beta=0.5$ & $\beta=0.6$ & $\beta=0.7$ & $\beta=0.8$ & $\beta=0.9$ & $\beta=1.0$ \\
 \midrule
4 Tasks & 62.0 & 61.9 & 62.5 & \textbf{62.8} & 62.1 & 61.6 & 60.7 & 61.0 & 60.6 & 61.0 \\
3 Tasks & 60.4 & 60.9 & 60.7 & 61.1 & \textbf{62.5} & \textbf{62.5} & 61.8 & 61.9 & 61.2 & 60.8 \\
2 Tasks & 60.0 & 60.3 & 60.4 & 61.2 & 60.6 & 61.9 & 61.7 & 61.9 & \textbf{62.2} & 62.0 \\
1 Task  & 60.1 & 60.4& 59.8 & 60.7 & 60.3 & 60.9 & 61.5 & 61.3 & 61.6 & \textbf{61.7} \\
\midrule
\multirow{3}{*}{4 Task} & \multicolumn{10}{c}{Ablation on the Selection of $\alpha$} \\
 & $\alpha=0.1$ & $\alpha=0.2$ & $\alpha=0.3$ & $\alpha=0.4$ & $\alpha=0.5$ & $\alpha=0.6$ & $\alpha=0.7$ & $\alpha=0.8$ & $\alpha=0.9$ & $\alpha=1.0$ \\
\cmidrule{2-11}
  & 55.7 & 57.6 & 59.2 & 61.1 & 59.5 & 60.2 & 61.1 & 61.0 & \textbf{62.8} & 59.3 \\
\bottomrule
\end{tabular}}
\end{table*}

\textbf{Ablation of $\beta$ and $\alpha$ in Sparse Vector Updating}\label{app:beta}
The detailed ablations regarding $\beta$ and $\alpha$ in sparse vector updating are presented in Table~\ref{ablation_beta}. In these experiments, we consistently use the reasoning segmentation task and employ the corresponding oIoU for evaluation. For instance, two tasks involved refer to reasoning segmentation and vanilla segmentation, while only one task involved refers to reasoning segmentation only. From the results, we observe that setting $\beta$ to 0.4 yields the best performance when all four tasks are involved. For scenarios involving three tasks, the optimal performance is achieved with $\beta$ set to 0.5 or 0.6. Finally, when there are only one or two tasks, it is more beneficial to activate a higher proportion of neurons, with $\beta$ set to 0.9 or 1.0. For the selection of $\alpha$ in EMA, we find that performance is optimal when $\alpha$ is set to 0.9, suggesting that cross-layer updating is more stable at this value. Conversely, smaller $\alpha$ values result in a dramatic performance drop, likely due to numerical instability in the cross-layer guidance updates. As a result, we set $\alpha$ to 0.9 in guidance signal updating.

\begin{table*}[t]
\centering
\caption{The ablation of activation patterns for sparse vector. In the experiments, all of four tasks are employed.}
\label{ablation_active}
\resizebox{\textwidth}{!}{
\begin{tabular}{c|cccccccccc}
\toprule
\multirow{2}{*}{Activation Pattern} & \multicolumn{10}{c}{oIoU of Reasoning Segmentation on MMUD} \\
 & $\beta=0.1$ & $\beta=0.2$ & $\beta=0.3$ & $\beta=0.4$ & $\beta=0.5$ & $\beta=0.6$ & $\beta=0.7$ & $\beta=0.8$ & $\beta=0.9$ & $\beta=1.0$ \\
 \midrule
Gaussian Random     & 62.0 & 61.9 & 62.5 & \textbf{62.8} & 62.1 & 61.6 & 60.7 & 61.0 & 60.6 & 61.0 \\
Top-$2\beta$ Random & 61.8 & 62.0 & 61.9 & \textbf{62.9} & 62.7 & 61.2 & 61.0 & 60.9 & 60.5 & 59.7 \\
Level Random        & 61.8 & 61.7 & 62.2 & \textbf{62.6} &\textbf{62.6}& 62.0 & 60.5 & 61.0 & 60.8 & 59.6 \\
Distribution Random & 62.1 & 60.9 & \textbf{62.5} & 62.4 & 60.6 & 59.8 & 59.7 & 60.0 & 59.6 & 60.1 \\
\bottomrule
\end{tabular}}
\end{table*}

\textbf{Ablation of Activation Patterns for Sparse Vector $Z_{k}^u$}\label{app:active}
Although we have conducted experiments on the activation rate $\beta$ of the sparse vector $Z_{k}^u$ (Appendix\ref{app:beta}), the activation pattern itself is still worth exploring. In designing our activation strategy, we aimed to activate different neurons according to different tasks and allow the number of activations to fluctuate slightly, mirroring the human thinking process. The simplest and most intuitive approach is to predefine an activation rate and allow it to fluctuate according to a Gaussian distribution, as illustrated in Section~\ref{sec:tuning}. However, other patterns can also adhere to this activation principle. Below are additional activation strategies explored in our experiments (assuming we need to activate $\beta\%$ of neurons and the fluctuation rate is $f\%$, where $f < \beta$):
\begin{itemize}
\item Top-$2\beta$ with Random Activation: First, we select the top $2\beta\% \pm 2f\%$ of neurons, then randomly activate $\beta\% \pm f\%$ of neurons within this selected group.
\item Level Random Activation: We sort and divide all neurons into 10 levels. In each level, we randomly activate $0.1\beta\% \pm f\%$ of neurons based on task instructions.
\item Distribution Random Activation: We define a hyperparameter $m$, then activate the $\beta\% \pm f\%$ of neurons closest to $m$ (in this approach, we set $m$ to 0 or 1).
\end{itemize}
\noindent The corresponding results for reasoning segmentation are shown in Table~\ref{ablation_active}. Surprisingly, we found that the specific activation strategy had little effect on the results (except the last distribution random activation strategy). Instead, the activation ratio $\beta\%$ had a more significant impact on the final outcomes, indicating that the model can consistently find the neurons it needs. Therefore, to keep the paper concise and easy to understand, we employ the simplest activation strategy in Section~\ref{sec:tuning}.

\section{Conclusion}
In this paper, we present a novel unified multimodal multitask learning framework with a new tuning strategy called neural tuning. Under this framework, we unify tasks using an all-in-token approach, which enhances scalability by facilitating the integration of additional modalities or tasks. Additionally, we emulate human cognitive processes through sparse distributed representation, activating specific neurons for different tasks. We evaluate our method across four tasks, including reasoning, segmentation, and text-to-image synthesis, demonstrating competitive performance compared to current state-of-the-art methods. To support further research in this area, we introduce the MMUD dataset, which provides a diverse set of annotations for various tasks. The tuned model weights will also be made publicly available to foster innovation and collaboration in this domain.

\subsection{Current Limitations}\label{limitation}
Despite the significant contributions of our work, certain limitations remain. Firstly, the current implementation does not incorporate acoustic modalities, which are critical for applications such as speech analysis and audio-visual reasoning. Secondly, while our framework has demonstrated strong performance across four tasks, its potential scalability to a broader range of tasks, including more than 10 simultaneous tasks, remains unexplored. The computational cost for such large-scale multitask training may become a concern, particularly for researchers with limited resources.

Furthermore, our framework currently employs LLaMA2 as the baseline LLM for experiments, leveraging both the 7B and 13B variants. While LLaMA2 is a strong and widely used model, it is not the most recent state-of-the-art in large language models. We acknowledge that adopting newer and more advanced LLMs could further enhance performance across tasks, particularly in scenarios demanding greater reasoning and generative capabilities. However, our experiments on multiple scales of LLaMA2 demonstrate that larger models consistently improve results, underscoring the scalability and generalizability of our approach. 

Additionally, the potential effects of task token overlap or interference were not a major concern in our experiments with four tasks using 128-dimensional sparse vectors. However, as the number of tasks increases, higher-dimensional sparse vectors may be required to mitigate interference, representing a theoretical aspect that warrants further investigation. Furthermore, the dependency on decoder choice for tasks like image synthesis was only partially explored in this work, as we evaluated VQGAN and DALL-E2 due to time constraints. Exploring alternative decoders could reveal additional insights into the generalizability of our framework.

\subsection{Future Research}
Our contributions lay a robust foundation for future advancements in multimodal multitask learning. The proposed MMUD dataset provides a valuable benchmark for evaluating such frameworks, and the release of model weights ensures reproducibility and facilitates adaptation across diverse applications. By leveraging a cognitive-inspired sparse task network and emphasizing parameter efficiency, our work highlights the intersection of neuroscience principles and machine learning methodologies.

Looking ahead, we aim to address the aforementioned limitations. This includes integrating acoustic modalities, extending the framework to accommodate tasks across diverse domains such as medical diagnostics, autonomous systems, and creative content generation, and conducting more comprehensive evaluations on public benchmarks. We also plan to optimize the framework to enhance its computational efficiency, ensuring accessibility for a wider range of researchers. Finally, we will also explore integrating the framework with cutting-edge LLMs, such as QWEN 3 or other advanced models, to fully exploit the latest advancements in multimodal understanding and multitask learning. These efforts will strengthen the robustness, scalability, and impact of our framework, further advancing the frontiers of multimodal multitask learning.

\bibliographystyle{elsarticle-num} 
\bibliography{references}

\appendix
\section{Data Filtration and Annotation Guidelines}
\label{app1}

To ensure the quality and suitability of the MMUD dataset for multimodal understanding, we employed a rigorous data filtration and annotation process following the initial data generation via GPT-4. Below, we outline the steps taken to refine the dataset and address potential issues in the generated content.

\subsection{Data Filtration Process}
After generating initial annotations using GPT-4, we manually reviewed and filtered the samples based on the following criteria:
\begin{itemize}
    \item Meaningfulness: Samples were excluded if the generated content was nonsensical or lacked coherence (for example: a simple response \textit{Yes, I know that answer.}).
    \item Length Appropriateness: Samples were filtered out if the length of the generated content was excessively long (longer than 500 words) or too short (shorter than 10 words) to provide meaningful multimodal understanding.
    \item Image Reference Dependency: Samples were discarded if the answers could be directly inferred from the questions without requiring reference to the associated images.
\end{itemize}

\subsection{Enhancement for Complex Reasoning}
To improve the dataset’s capability to support complex reasoning tasks, we introduced object tokens ($<$OBJ-i$>$) in the generated answers. Each $<$OBJ-i$>$ token corresponds to a specific object in the image, enabling the models to better align textual descriptions with visual content.

For example, for a question such as:

\textit{Can you judge the relationship between the two people in the front based on what happened in the picture?}

The initial answer was:

\textit{In this image, one of the women in the front is cutting a pizza, and a little girl is next to her...}

We manually inserted object tokens to provide a more structured and precise representation:

\textit{In this image, one of the women in the front $<$OBJ-1$>$ is cutting a pizza $<$OBJ-2$>$, and a little girl $<$OBJ-3$>$ is next to her...}

\begin{figure*}[t]
\centering
\begin{subfigure}{0.32\linewidth}
    \centering
    \includegraphics[width=1.0\textwidth]{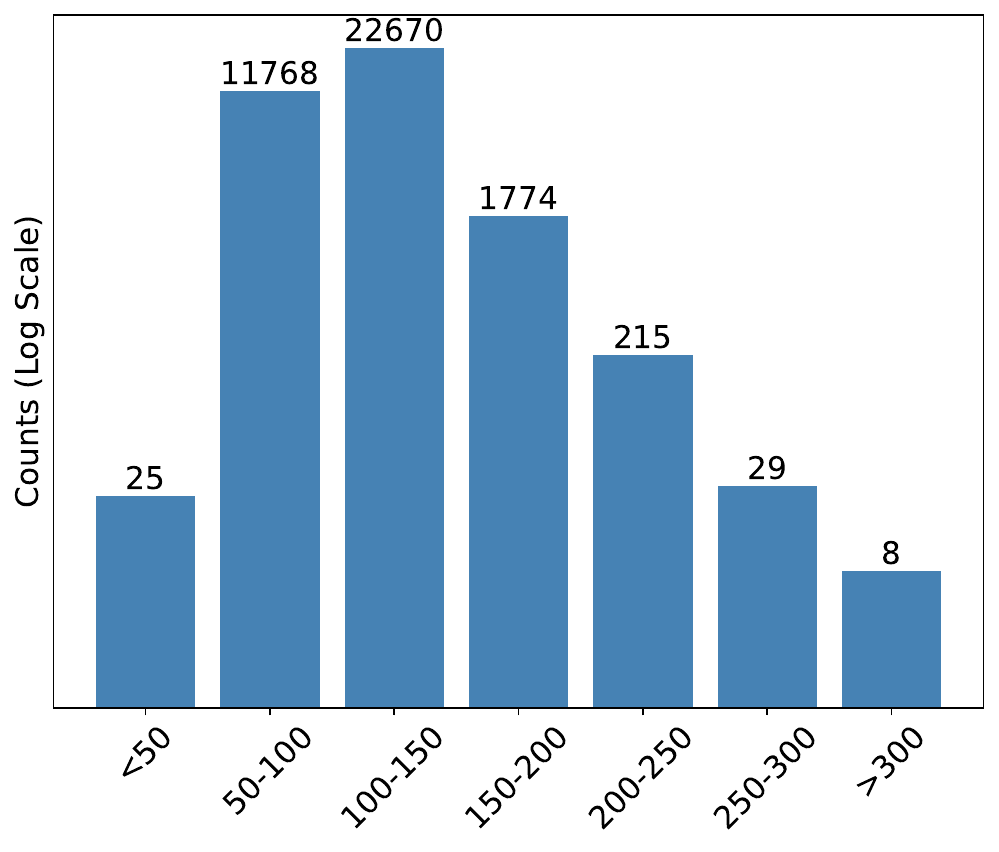}
    \caption{The caption length distribution in MMUD.}
    \label{fig:caption_length}
\end{subfigure}
\hfill
\begin{subfigure}{0.32\linewidth}
    \centering
    \includegraphics[width=1.0\textwidth]{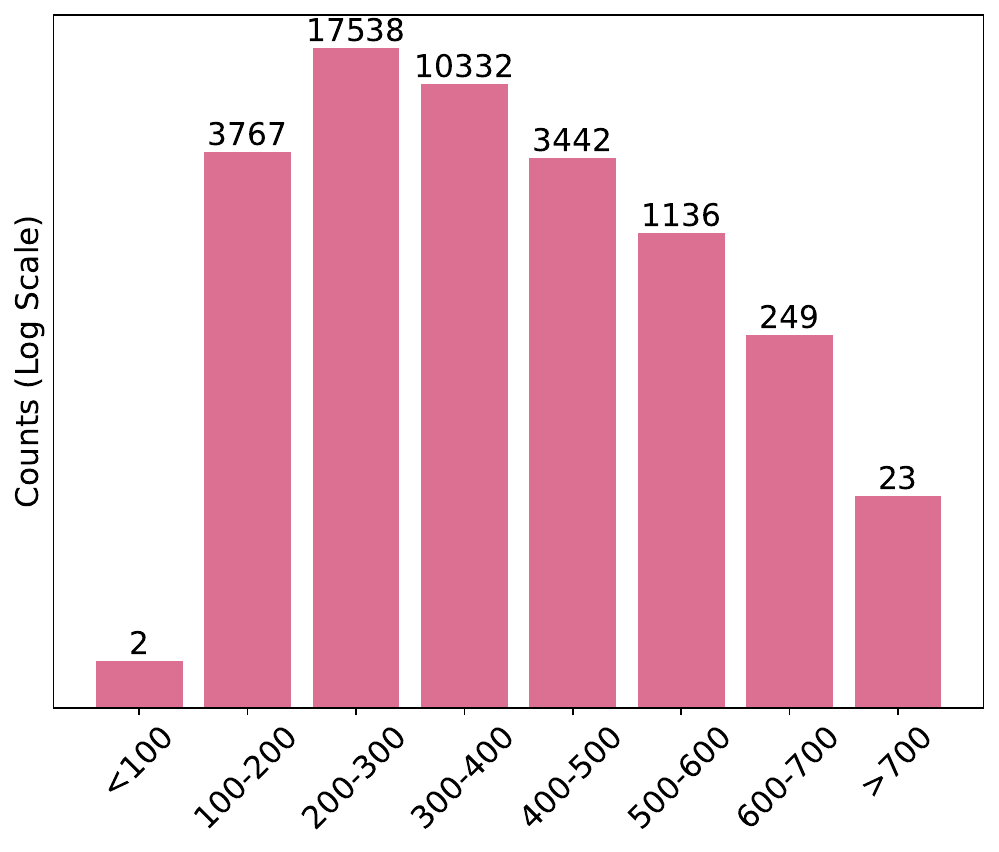}
    \caption{The reasoning QA pair length distribution in MMUD.}
    \label{fig:reason_length}
\end{subfigure}
\hfill
\begin{subfigure}{0.32\linewidth}
    \centering
    \includegraphics[width=1.0\textwidth]{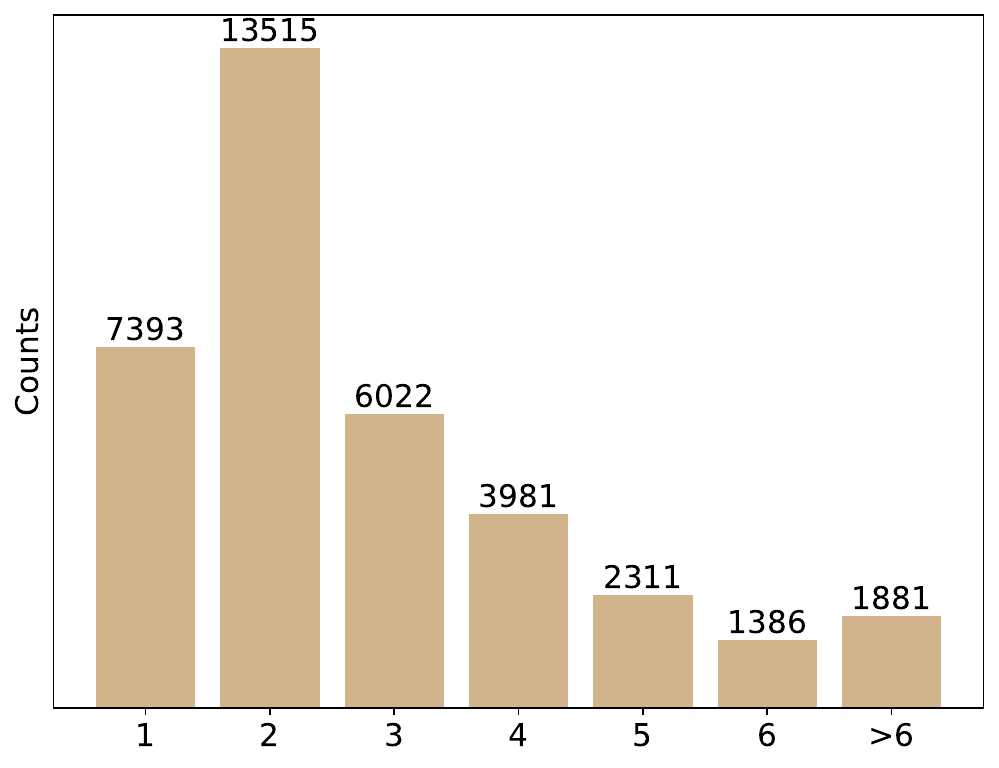}
    \caption{The object counts per sample in MMUD.}
    \label{fig:object_length}
\end{subfigure}
\caption{The distribution of caption length, reasoning question-answer pair length, and object counts for each sample in MMUD.}
\end{figure*}

\begin{table*}[t]
    \centering
    \caption{The comparison of neural tuning, LoRA, and prefix tuning on RefCOCO, RefCOCO+, and RefCOCOg. The metric used in the table is oIoU.}
    \label{result_lora_compare}
    \resizebox{1.0\textwidth}{!}{
    \begin{tabular}{cccccccccc}
    \toprule
    \multirow{2}{*}{Method} & \multirow{2}{*}{TFLOPs} & \multicolumn{3}{c}{RefCOCO} & \multicolumn{3}{c}{RefCOCO+} & \multicolumn{2}{c}{RefCOCOg(U)} \\
    &  & Val & TestA & TestB & Valid & TestA & TestB & Valid & Test \\
    \midrule
    Prefix Tuning-7B & 3.32 & 70.3 & 72.7 & 65.0 & 61.1 & 67.8 & 55.7 & 67.1 & 66.2 \\
    LoRA-7B          & 2.38 & 72.1 & 74.2 & 67.3 & 64.2 & 69.9 & 56.4 & 68.0 & 67.1 \\
    NT-7B(Ours) & 2.47 & \textbf{74.2} & \textbf{76.7}& \textbf{68.0} & \textbf{66.3} & \textbf{71.2} & \textbf{58.1} & \textbf{70.3} & \textbf{70.2 }\\
    \bottomrule 
    \end{tabular}}
\end{table*}
\subsection{Object Annotations and Masks}
To ensure the quality and utility of the MMUD dataset, we manually annotated all 36,582 samples, with each sample containing descriptions of 2.8 objects in average. For each object, we inserted an $<$OBJ-i$>$ tag after its corresponding description to facilitate multimodal reasoning and segmentation tasks. During the process, we ensured that each tag corresponds to its respective mask in the RefCOCO series datasets. This annotation process was completed over a period of two months by three dedicated annotators. While labor-intensive, this effort was crucial for creating a high-quality dataset that aligns with the goals of multimodal multitask learning research. The detailed annotations provide a robust foundation for various tasks, ensuring that the dataset is both comprehensive and reproducible for the research community.

After data annotation, to provide a comprehensive overview of the MMUD dataset, we present key metrics that highlight its diversity and richness. Figure~\ref{fig:caption_length} illustrates the distribution of caption lengths, showing that most captions fall within the range of 50 to 200 tokens, reflecting the dataset's emphasis on detailed and descriptive annotations. Similarly, Figure~\ref{fig:reason_length} depicts the length distribution of reasoning question-answer pairs, with the majority spanning 100 to 400 tokens, underscoring the complexity and depth of reasoning required for these tasks. Lastly, as shown in Figure~\ref{fig:object_length}, the object counts per sample reveal that most samples contain fewer than three objects, with two objects being the most common. This distribution aligns with the design of the dataset to balance between simplicity and complexity, ensuring practical usability across a range of multimodal applications.

\section{Comparison of Neural Tuning with LoRA and Prefix Tuning}
We evaluated the parameter efficiency and performance of our neural tuning mechanism against common alternatives, including Low-Rank Adaptation (LoRA)~\cite{hu2021lora} and prefix tuning~\cite{li2021prefix}. These experiments on vanilla referring segmentation tasks. The results are summarized in Table~\ref{result_lora_compare}. Neural tuning demonstrates similar TFLOPs to LoRA, indicating comparable computational efficiency, while consistently achieving superior performance across all evaluated metrics. Specifically, neural tuning achieves an oIoU of 74.2\%, compared to 72.1\% for LoRA and 70.3\% for prefix tuning.

These results highlight neural tuning as a balanced and scalable solution for multitask and multimodal learning. Its advantage stems from dynamically allocating computation while maintaining lightweight operations, surpassing traditional tuning strategies in both performance and efficiency.

%The results indicate that neural tuning balances computational efficiency and performance effectively, making it a practical and scalable approach for multitask and multimodal learning. This improvement can be attributed to the mechanism’s ability to allocate computational resources dynamically while maintaining lightweight operations, showcasing its advantage over traditional tuning methods.

\begin{table*}[t]
\centering
\caption{The multitask interference analysis for tasks involved when activating all neurons in sparse vectors.}
\label{result_interference}

\resizebox{1.0\linewidth}{!}{
\begin{tabular}{c|ccc}
\toprule
Main Task & Additional Task & Performance & $\Delta$ \\
\midrule
\multirow{4}{*}{Referring Segmentation (oIoU on RefCOCO TestA Set)} & - & 75.4 & - \\
 & Reasoning Segmentation & 76.0 & +0.6\% \\
 & Imgae Captioning & 75.4 & +0.0\%  \\
 & Text-to-Image Generation & 74.9 & -1.5\%  \\
 \midrule
\multirow{4}{*}{Reasoning Segmentation (oIoU on MMUD Test Set)} & - & 61.7 & - \\
 & Referring Segmentation & 62.0 & +0.3\% \\
 & Imgae Captioning & 61.1 & -0.6\% \\
 & Text-to-Image Generation & 60.4 & -1.3\%\\
 \midrule
 \multirow{4}{*}{Imgae Captioning (BLEU-4 on MMUD Test Set)} & - & 44.1 & - \\
    & Reasoning Segmentation & 44.0 & -0.1\\
    & Referring Segmentation & 49.7 & -0.4\\
    & Text-to-Image Generation & 42.8 & -1.3\\
 \midrule
\multirow{4}{*}{Text-to-Image Generation (FID($\uparrow$) on MMUD Test Set)} & - & 11.9 & - \\
    & Reasoning Segmentation & 12.8 & +0.9 \\
    & Referring Segmentation & 13.0 & +1.1 \\
    & Image Captioning & 12.2 & +0.3 \\
 \bottomrule
\end{tabular}}
\end{table*}

\section{Multitask Interference Analysis}
%This section details experiments conducted to analyze task interference when all neurons are activated during multitask training. The goal was to understand how tasks interact and influence one another within the unified training framework.

%We evaluated four tasks: referring segmentation, reasoning segmentation, text-to-image generation, and captioning, focusing on performance when all neurons were activated without sparsity constraints. As we could draw from the results (Table~\ref{result_interference}), tasks with high semantic correlation, such as referring segmentation and reasoning segmentation, demonstrated mutual benefits. Shared features like spatial reasoning and object localization resulted in oIoU improvements of 0.6\% and 0.3\%, respectively. However, tasks with divergent objectives showed negative interactions. For example, reasoning segmentation and text-to-image generation experienced performance drops due to conflicting neuron activation patterns, with reasoning segmentation’s oIoU decreasing by 0.3\% and text-to-image generation’s FID score increasing by 0.3. Captioning, relying on distinct neuron subsets, exhibited minimal interaction with other tasks, though slight fluctuations indicated mild interference.

This section presents experiments analyzing task interference when all neurons are activated during multitask training. The objective was to examine how tasks interact within a unified framework.

We evaluated four tasks: referring segmentation, reasoning segmentation, text-to-image generation, and captioning, under full neuron activation without sparsity. As shown in Table~\ref{result_interference}, semantically related tasks—such as referring and reasoning segmentation—benefited from shared features like spatial reasoning and object localization, yielding oIoU gains of 0.6\% and 0.3\%, respectively. In contrast, tasks with divergent goals exhibited negative interference. Reasoning segmentation and text-to-image generation showed conflicting activation patterns, resulting in a 0.3\% drop in oIoU and a 0.3 increase in FID, respectively. Captioning, relying on more distinct neuron subsets, showed minimal interaction, though slight performance fluctuations suggest mild cross-task interference.

%To address these issues, we adopted a sparse neuron activation strategy inspired by neuroscience, selectively activating neurons for each task. This mitigated negative interactions while preserving positive correlations, as shown by the performance stabilization across all tasks. The sparsity parameter $\beta$, empirically set to 0.4 (see Section~\ref{result_ablation} and Table~\ref{ablation_beta}), effectively eliminated task interference, allowing for consistent performance while maintaining multitask efficiency.

%This analysis highlights the importance of task-specific strategies in multitask frameworks. By reducing neuron overlap, our neural tuning approach balances task interactions, enhancing both individual and overall performance.

To address task interference, we adopted a neuroscience-inspired sparse neuron activation strategy that selectively activates neurons per task. This approach mitigates negative interactions while preserving beneficial correlations, leading to stable performance across tasks. The sparsity parameter $\beta$, empirically set to 0.4 (see Section~\ref{result_ablation} and Table~\ref{ablation_beta}), effectively reduces overlap and supports consistent, efficient multitask learning.

These findings underscore the value of task-specific strategies in multitask frameworks. By minimizing neuron overlap, our neural tuning method strikes a balance between task isolation and shared learning, boosting both individual and overall performance.

\end{document}